\documentclass{article}

\usepackage{microtype}
\usepackage{graphicx}
\usepackage{subfigure}
\usepackage{booktabs} % for professional tables

\usepackage{hyperref}

\usepackage[accepted]{icml2024}

\usepackage{amsmath}
\usepackage{amssymb}
\usepackage{mathtools}
\usepackage{amsthm}
\usepackage{amsmath}
\usepackage{algorithmic}
\usepackage{algorithm}
\usepackage{enumitem} 
\usepackage[marginal]{footmisc}
\usepackage{rotating}
\usepackage{textcomp}
\usepackage{textcomp}
\usepackage{stfloats}
\usepackage{verbatim}
\usepackage{graphicx}
\usepackage{bm}

\usepackage[capitalize,noabbrev]{cleveref}

\theoremstyle{plain}

\theoremstyle{definition}

\theoremstyle{remark}

\usepackage[textsize=tiny]{todonotes}

\begin{document}

\twocolumn[
\icmltitle{NeuralOOD: Improving Out-of-Distribution Generalization Performance with Brain-machine Fusion Learning Framework}



\icmlsetsymbol{equal}{*}

\begin{icmlauthorlist}
\icmlauthor{Shuangchen Zhao}{equal,cnu,casia}
\icmlauthor{Changde Du}{equal,casia,casia2,ucas}
\icmlauthor{Hui Li}{cnu}
\icmlauthor{Huiguang He}{casia,casia2,ucas}
\end{icmlauthorlist}

\icmlaffiliation{cnu}{Faculty of Education, Capital Normal University, Beijing, China}
\icmlaffiliation{casia}{Laboratory of Brain Atlas and Brain-inspired Intelligence, Institute of Automation, Chinese Academy of Sciences, Beijing, China}
\icmlaffiliation{casia2}{Key Laboratory of Brain Cognition and Brain-inspired Intelligence Technology, Chinese Academy of Sciences, Beijing, China}
\icmlaffiliation{ucas}{School of Artificial Intelligence, University of Chinese Academy of Sciences, Beijing, China}

\icmlcorrespondingauthor{Hui Li}{lihui@cnu.edu.cn}
\icmlcorrespondingauthor{Huiguang He}{huiguang.he@ia.ac.cn}

%\icmlkeywords{Machine Learning, ICML}

\vskip 0.3in
]



\printAffiliationsAndNotice{\icmlEqualContribution} % otherwise use the standard text.

\begin{abstract}
 Deep Neural Networks (DNNs) have demonstrated exceptional recognition capabilities in traditional computer vision (CV) tasks. However, existing CV models often suffer a significant decrease in accuracy when confronted with out-of-distribution (OOD) data. In contrast to these DNN models, human can maintain a consistently low error rate when facing OOD scenes, partly attributed to the rich prior cognitive  knowledge stored in the human brain. Previous OOD generalization researches only focus on the single modal, overlooking the advantages of multimodal learning method. In this paper, we utilize the multimodal learning method to improve the OOD generalization and propose a novel Brain-machine Fusion Learning (BMFL) framework. We adopt the cross-attention mechanism to fuse the visual knowledge from CV model and prior cognitive knowledge from the human brain. Specially, we employ a pre-trained visual neural encoding model to predict the functional Magnetic Resonance Imaging (fMRI) from visual features which eliminates the need for the fMRI data collection and pre-processing, effectively reduces the workload associated with conventional BMFL methods. Furthermore, we construct a brain transformer to facilitate the extraction of knowledge inside the fMRI data. Moreover, we introduce the Pearson correlation coefficient maximization regularization method into the training process, which improves the fusion capability with better constrains. Our model outperforms the DINOv2 and baseline models on the ImageNet-1k validation dataset as well as six curated OOD datasets, showcasing its superior performance in diverse scenarios.
\end{abstract}

\section{Introduction}
\label{introduction}

Deep learning \cite{lecun2015deep} has achieved remarkable success in various fields such as Computer Vision (CV) and Natural Language Processing (NLP), primarily due to the availability of high-quality, large-scale, well-annotated datasets (e.g., ImageNet-1k dataset \cite{imagenet}) and the development of well-designed model architectures. In recent years, numerous CV models have continually set new benchmarks in the ImageNet classification task, and the paradigm has also transitioned from Convolutional Neural Networks (CNNs) \cite{cnn} to Vision Transformers (ViTs) \cite{vit}. These Deep Neural Networks (DNNs) based on the Empirical Risk Minimization (ERM) optimization method (e.g., gradient descent) have demonstrated impressive image recognition capabilities, surpassing human performance under the assumption of independent and identically distribution (${i.i.d.}$) \cite{krizhevsky2012imagenet}. However, DNNs' effectiveness tends to diminish when confronted with extreme environments that deviate from the ${i.i.d.}$ assumption, such as out-of-distribution (OOD) samples. In the real-world applications, the decrease in accuracy caused by the ubiquitous distribution shift phenomenon poses potential risks in applications such as autonomous driving, medical diagnosis, and security systems, leading to a sharp increase in model unreliability. This limitation underscores the need to develop more robust models that can maintain high performance across a broader range of conditions and data distributions, thereby reducing the catastrophic failures caused by domain shift.

In order to address the aforementioned shortcomings, domain generalization (DG), also known as OOD generalization, was spawned as a critical area of the transfer learning research. OOD generalization aims to train a model not only  excel on ${i.i.d.}$ test dataset but also generalized to unknown distributions \cite{arjovsky2021distribution,dgsurveymsra}. This issue remains a central focus in both academic and industrial research due to its significant implications \cite{dgsurveymsra}. Several works have attempted to improve the OOD generalization performance through unsupervised representation learning methods, supervised models and optimization methods \cite{liu2021towards}. Although these methods have made significant improvements, their effectiveness remains substantially inferior to the human performance. Human performs well in extreme environment \cite{naturebmf}, while traditional CV models are unable to achieve the same level of generalization ability \cite{cvprbmf}. There are some interesting commonalities between the DNNs and human brain. Although previous works have indicated that the occipital region of the human brain is primarily responsible for visual information processing \cite{grill1998sequence,Jacobson2008,neuron,de2008perceived}, the refine visual processing is still vague. Consequently, the interpretability of DNNs is also uncertain. DNNs are designed to imitate the structure of human brain neurons, with the convolutional layers of CNNs also designed to mimic the visual information processing of human brain. DNNs and human brain shows their similarity on both structure and function.

\begin{figure}
	\centering
	\includegraphics[width=\linewidth]{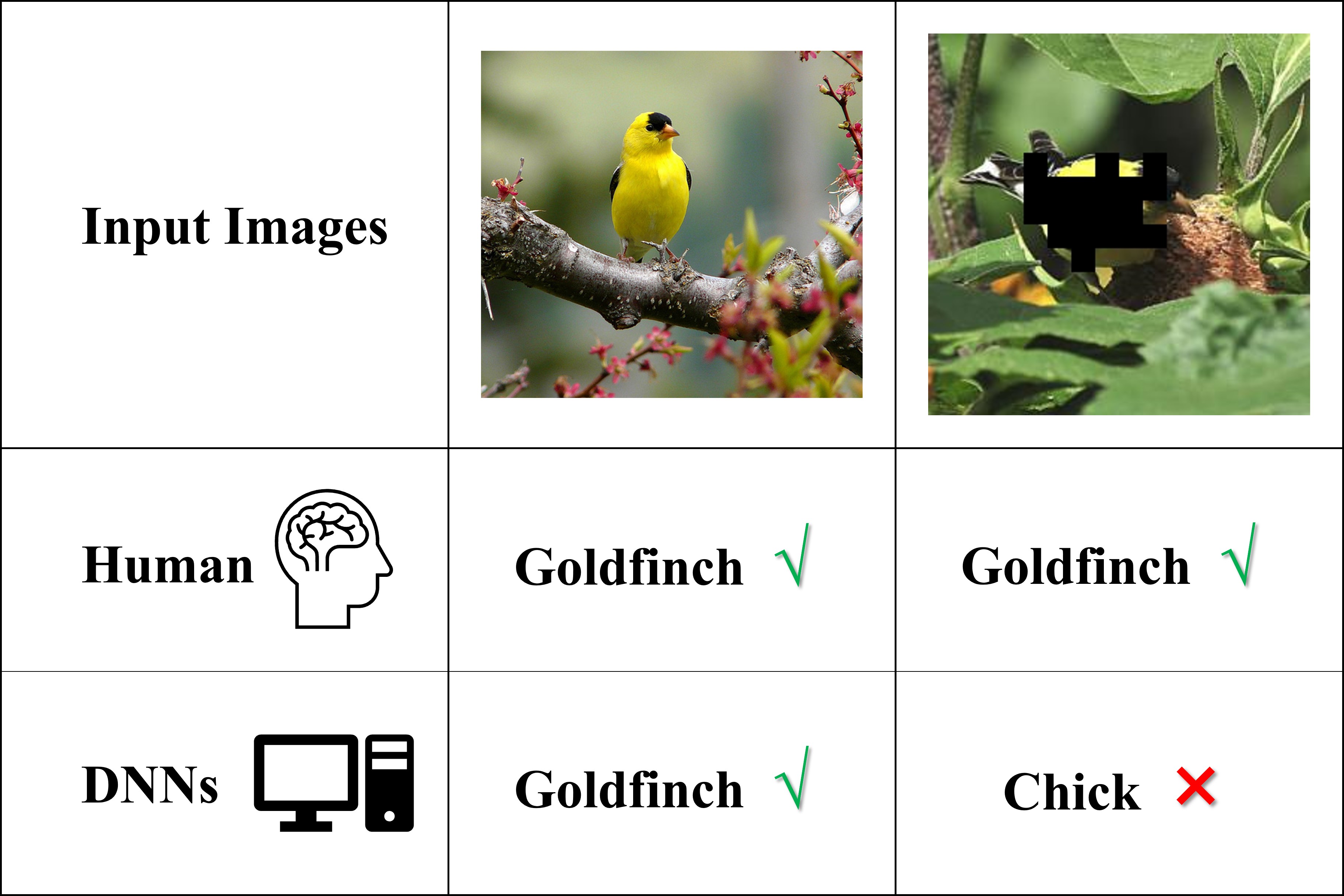}
	\caption{DNNs can achieve or even surpass human recognition capabilities when presented with common data. However, their recognition performance significantly deteriorates when faced with OOD data, in contrast to the resilient recognition abilities exhibited by humans in such scenarios.}
	\label{fig:illustrate}
\end{figure}

As the most intricate and efficient computation platform in the world, the human brain can easily extract abstract features from complex items in natural scenarios with minimal energy consumption \cite{naturebmf}. Although DNNs, which are designed by imitating the structure of neurons in the human brain, perform well in many situations, human visual recognition ability is more reliable than DNNs \cite{du2023decoding}, especially in extreme environments. For example, when we are driving in the midnight, it's easy for us to identify different items and make proper decisions while DNNs can not. We describe the different distribution between the normal and extreme environments as the domain shift phenomenon \cite{learningdataset}. Human can overcome the domain shift phenomenon but DNNs struggle to generalize to different distributions. The disparity in performance between human and conventional CV models in OOD generalization may stem from the substantial prior knowledge stored in the human brain. 
%As the most intricate and efficient computation platform, the human brain can easily extract abstract features from complex elements in the natural world with minimal energy consumption \cite{naturebmf}. DNNs are also designed to imitate the structure of neurons in the human brain 
%IN CONTRAST, DNNS STRUGGLE TO EFFECTIVELY ADDRESS DOMAIN SHIFT CHALLENGES ARISING FROM DIFFERENT DISTRIBUTIONS \CITE{LEARNINGDATASET}. 
Some researchers have designed the brain-inspired models to imitate the physiological structure of human brain, successfully improving the relevant task performance from the architecture level \cite{10175605,10123695,9418554,9511841,KubiliusSchrimpf2019CORnet}. While other researches found that prior knowledge in human brain can help human accumulate new knowledge \cite{brod2013influence}. Several works \cite{futac,du2023decoding,liu2023brain,9097411,naturebmf,cvprbmf} have also successfully improve the performance of traditional model through brain-machine fusion learning method. Comparing with brain-inspired models, brain-machine fusion models can fuse cognitive knowledge with visual knowledge, using multimodal information to improve the perception ability of DNNs. 

Inspired by the aforementioned methods, we propose a novel Brain-machine Fusion Learning (BMFL) framework to further improve the OOD generalization ability of CV models and subsequently improve its robustness in real-world scenarios. The proposed model mainly focuses on sharing the cognitive knowledge with DNN visual features, letting prior cognitive  knowledge to guide the perception process of CV models. In the proposed model, we utilize functional Magnetic Resonance Imaging (fMRI) data as the brain modality because of its high spatial resolution, which makes it more suitable for contributing to the perception process in CV models. Previous neural decoding works usually utilize public brain datasets such as Natural Scenes Dataset (NSD) \cite{nsd}, Generic Object Decoding (GOD) dataset \cite{GOD}, Deep Image Reconstruction (DIR) dataset \cite{DIR}, and Human Connectome Project (HCP) dataset \cite{HCP}, etc.. The limitations of public brain fMRI  datasets are obvious. Firstly, same as the other methods of recording brain signals, fMRI exhibits significant inter-subject variability. Moreover, the data volume from a single subject is commonly limited by the various types of fatigue experienced by the human brain, including motor fatigue \cite{van2007effects} and chronic fatigue \cite{de2004neural}, during the fMRI data acquisition process. Then, the collection and processing of human brain visual fMRI data can be a highly expensive endeavor. To address the aforementioned limitations, we employ a pre-trained fMRI prediction model to generate large-scale fMRI data with no extra cost. Our work introduces a new idea on OOD generalization tasks, the main contributions of this paper can be summarized as follows:
%However, on the one hand, data plays an important role in model's generalization ability, existing fMRI-image paired datasets are far from our requirements. On the other hand, collecting and processing human brain visual fMRI data can be a highly expensive endeavor. Thus, generating brain visual fMRI from the  pre-trained visual neural encoding model is more economical and time-saving. Our work provides a new idea on OOD generalization tasks, we can further apply BMF learning method to brain-computer interface (BCI). Our main contributions can be summarized as follows:

\begin{itemize}
	
	\item We propose a novel BMFL framework that improves the OOD generalization capability of its CV backbone.
	
	\item The proposed BMFL framework extracts the prior cognitive knowledge contains in the human brain through the brain transformer module, and fuses the prior cognitive knowledge with the CV features by the cross-attention mechanism with the restriction of Pearson correlation coefficient maximization method.
	
	\item We have created and open-sourced two OOD datasets based on ImageNet-1k validation set \footnote{\url{ https://figshare.com/articles/dataset/BMF_datasets/26393533}}.
	
	\item Previous public OOD datasets have less practical value because they are from different sources (e.g., selected from different retrieval methods), whereas our datasets are from the same source with different distributions (e.g., low light and occluded images) that have high practical value in real-world applications.
	
\end{itemize}

\section{Related Work}
The proposed BMFL framework is a brain-guide multimodal model which efficiently improves the OOD generalization capability of its visual backbone. In this section, we review the development of the OOD generalization task first. Then we discuss the differences between brain-inspired models and brain-machine fusion models. In the end, we sum up the multimodal learning methods based on the cross-attention mechanism.

\subsection{Out-of-Distribution Generalization}
OOD generalization aims to train a model utilizing one or several in-domain (ID) source data that can generalize to the target unseen (i.e., OOD) domain(s) \cite{dgsurveymsra}. The main paradigm of OOD generalization can be broadly categorized into three types: data manipulation, representation learning, and learning strategy. Among these, representation learning, which improves the OOD generalization ability by learning the domain-invariant features, stands out as the most popular method. Some researchers propose an  Invariant Risk Minimization (IRM) \cite{IRM} method, which learns the domain-invariant features across different domains by minimizing the domain invariant penalty, while other researchers conduct a knowledge distillation method to learn the domain-invariant features representations \cite{10093034}.

Although IRM has demonstrated high OOD generalization ability, its shortcomings are also evident. On the one hand, IRM-based methods perform well on shallow networks, such as linear models, but they are prone to overfitting in deeper networks. On the other hand, IRM is sensitive to the environment partitioning, the effectiveness of IRM may be significantly reduced if the partitioning is not sufficiently reasonable. To address the overfitting issue in the  overparameterized networks, the SparseIRM method \cite{sparseirm} is proposed to introduce a global sparsity constraint into the IRM process. Futhermore, \cite{irmnips} utilizes additional auxiliary information as the prior knowledge to learn the domain-invariant features, thereby reducing the dependency on environment partitioning if IRM. Several works improve OOD generalization performance under the traditional ERM framework, \cite{nasood} tackles the two aforementioned issues by introducing the Neural Architecture Search method into the research of OOD generalization problems, \cite{9582738} uses conditional variational autoencoder (VAE) and gradient reversal layer to achieve high accuracy.

The above methods refine the OOD generalization ability by introducing auxiliary information or altering the training process, but they mainly focus only on the single image modality. Although \cite{irmnips} uses the auxiliary information, data from other modalities can contain richer information compared to the auxiliary information. Therefore, we propose a BMFL framework, utilizing multimodal data as well as multimodal learning method to improve the OOD generalization of DNNs. Comparing with the conventional IRM methods, the proposed BMFL framework can obtain higher OOD generalization accuracy with ERM optimization method that eliminates the dependency on environment partitioning.

\subsection{Brain-based Models}
Brain-based models can be summarized into two main categories: brain-inspired models and brain-machine fusion models. Spiking Neural Networks (SNNs) is a typical brain-inspired model, which is designed to imitate the biological neuron system. During the information transmission process, neurons emit spikes only when the membrane potential reaches the threshold value. Due to its elaborate design, SNNs can improve the computational efficiency while reducing the power consumption. Based on the above advantages and the strong bio-neurological basis, there are several works conduct their research based on the SNNs architecture \cite{10175605,10123695,9418554,9511841}. Recently, some researchers proposed a brain-score \cite{brainscore} to measure the similarity between brain visual streams and DNNs. DenseNet-169 \cite{huang2017densely}, CORnet-S \cite{KubiliusSchrimpf2019CORnet}, and ResNet-101 \cite{ResNet} are considered as the most brain-like models. Different from other models, CORnet-S uses shallow networks with only four layers which corresponds to the four brain visual areas and obtains high image  classification accuracy than the deeper models. Both SNNs and brain-like models belong to the structural improvement of DNNs.

Comparing with those brain-inspired models, the brain-machine fusion models can usually achieve higher performance in different tasks. Previous multimodal learning methods such as CLIP \cite{clip}, obtains a high classification accuracy and even demonstrates the impressive zero-shot classification capability. Comparing with previous multimodal models, the brain-machine fusion models introduce a more robust prior cognitive knowledge into the learning system, which will further improve the performance on variant tasks. The BraVL model \cite{du2023decoding} utilizes a  multimodal VAE (i.e., Mixture of Product of Experts) to fuse the brain, visual, language modalities. Restricted by evidence lower bound and mutual information maximization, the BraVL model demonstrates proficient performance in the zero-shot visual neural decoding task. Some researchers designed a VLD model \cite{HUANG2024102573} that introduces Bidirectional Gated Recurrent Unit into the brain visual neural decoding task and use multitask encoder to joint modeling the fMRI-task information. In the end, encoded features will be decoded from the unique task models. Palazzo \emph{et al.} proposed a siamese network to learn the multimodal joint representation of EEG and the image stimuli (i.e. Cortical-Visual Representations). It performs well in the EEG classification, image classification and saliency detection tasks \cite{9097411}. Liu \emph{et al.} proposed a brain-machine coupled learning method for facial emotion recognition (FER) task \cite{liu2023brain}. They utilized three channels (i.e. visual channel, cognitive channel and common channel) to extract visual knowledge, cognitive knowledge and brain-guide-visual knowledge. Their brain-machine coupled learning method obtains the SOTA performance of the FER classification task. Mean while, Quan \emph{et al.} utilized multimodal contrastive learning algorithm for brain-machine fusion learning \cite{QUAN2024102447}. Specifically, they align the EEG signal and its stimulus images via constractive learning method during the training process. The aligned frozen-weighted image encoder will be used as the classification task to train the SVM classifier. Some researchers successfully increased classification accuracy \cite{naturebmf} and model robustness \cite{cvprbmf} by introducing brain signals to DNNs. Fu \emph{et al.} used the representation similarity analysis to combine brain knowledge with CNN features, thereby improving the video emotion recognition accuracy \cite{futac}. Several works \cite{fmrieegbert} also introduce cognitive knowledge into the NLP research, using cognitive knowledge to improve the performance of NLP models in relevant NLP downstream tasks.

\subsection{Multimodal Learning Using Cross-attention Mechanism}
Since the remarkable success of the transformer architecture \cite{attention} in NLP \cite{BERT} and CV \cite{vit} domains, the advent of a unified framework has rendered multimodal learning architecturally viable. As an attention-based method, the cross-attention mechanism can fuse the information from different sequences or modalities. Thus, it performs well on the machine translation, multi-scale input and multimodal learning tasks. In the single-modal learning domain, some researchers introduces the cross-attention mechanism into the image classification task \cite{crossvit}. By integrating multi-scale inputs from ViTs of different sizes, it efficiently improves the classification performance of the conventional ViT models. In the multimodal learning domain, various Vision-Language Pre-training models \cite{albef,blip2} demonstrate their ability to comprehend the semantic information of the input images. They can answer questions posed by people based on the content of the images. Cross-attention mechanism is the key of its powerful cross-modal content comprehension ability. Through the joint representation modeling of images and text by cross-attention mechanism, the domain distribution of different modalities are aligned. Aside from its cross-modal comprehension capability, the cross-attention mechanism also demonstrates high cross-modal generation ability. Concurrently, some text-to-image generative models \cite{latentdiffusion,dalle} show the capability to generate high-quality images through few sentences via cross-attention mechanism. This cross-modalities approach has facilitated a deeper comprehension and synthesis of information across different modalities. Recent study \cite{crossood} proves that different model architectures exhibit varying performance on OOD generalization tasks, models with multiple attention layers or using cross-attention mechanism have a better OOD generalization performance.

\section{Methodology}

\subsection{Problem Definition and Framework Overview}
There are at least two distinct domains in the OOD generalization task (e.g., training domain $\mathcal D^{trn}$ and testing OOD domain $\mathcal D^{ood}$, etc.), our goal is to build a model $\mathcal F$ that has good performance on both $i.i.d$ test dataset from the training set and OOD testing dataset. During the training process, we optimize the model $\mathcal F$ by minimizing the loss function $\mathcal L$ with training set labels $\mathcal Y$ and the predicted labels $\mathcal{Y}_{pred}$. The optimization process can be defined as: 
\begin{equation}
	\label{eq:definition}
	\mathcal{F}_{o} = \arg\min \mathop{\mathbb{E}}\limits_{\mathcal X,\mathcal Y \sim \mathcal D^{trn}} [\mathcal{L}(\mathcal{F}(\mathcal X),\mathcal Y)]
\end{equation}
where $\mathcal X$ denotes the training set data while $\mathcal{F}_{o}$ denotes the model we aim to find.

Let $D_{num}$ denotes different datasets and $\mathcal D^{num}$ refers to the variant distributions, where $D_{num}=\{(\mathcal X_{num}, \mathcal Y_{num})\}$. Assume that there are n datasets with n-1 different distributions ($D_{trn}, D_{val} \in \mathcal{D}^{trn}$) in the OOD generalization task. Then $\mathcal D^{OOD}=\{\mathcal D^{1}, \mathcal D^{2}, …, \mathcal D^{n-1}|D_1 \in \mathcal D^1, D_2 \in \mathcal D^2, …, D_{n-1} \in \mathcal D^{n-1}\}$. Specifically, $D_{val}$ is $i.i.d.$ with $D_{trn}$, $\mathcal X_{trn} \cap \mathcal X_{val} = \phi, \mathcal Y_{trn}=\mathcal Y_{val}$. Moreover,  $X_{OOD}=\{D^{1}, D^{2}, …, D^{n-1}\}$ are mutually independent, reflecting distinct OOD scenarios. This independence distribution of different domains ensures that our BMFL framework can generalize from seen domain (i.e., $\mathcal{D}^{trn}$) to unseen OOD domains (e.g., $\mathcal D^1$, $\mathcal D^2$, etc.).

%In this paper, we use 7 test datasets with different distributions to test the OOD generalization capability for baselines as well as the proposed BMF model. We assume that $D_x$ denotes different datasets and $\mathcal D^x$ refers to variant distributions, then, $D_{trn}, D_{val} \in \mathcal{D}^{trn}$, where $D_{trn} \cap D_{val} = \phi$. As mentioned before, there are 7 variant datasets (i.e., $D_1$, $D_2$, …, $D_7$) and its corresponding distributions (i.e., $\mathcal D^1$, $\mathcal D^2$, …, $\mathcal D^7$). Moreover,  $\mathcal D^1$, $\mathcal D^2$, …, $\mathcal D^7$ are mutually independent, reflecting distinct OOD scenarios. This independence distribution of different domains ensures that our BMF model can generalize from seen domain (i.e., $\mathcal{D}^{trn}$) to unseen OOD domains (e.g., $\mathcal D^1$, $\mathcal D^2$, etc.).

\begin{figure*}[htbp!]
	\centering
	\includegraphics[width=\linewidth]{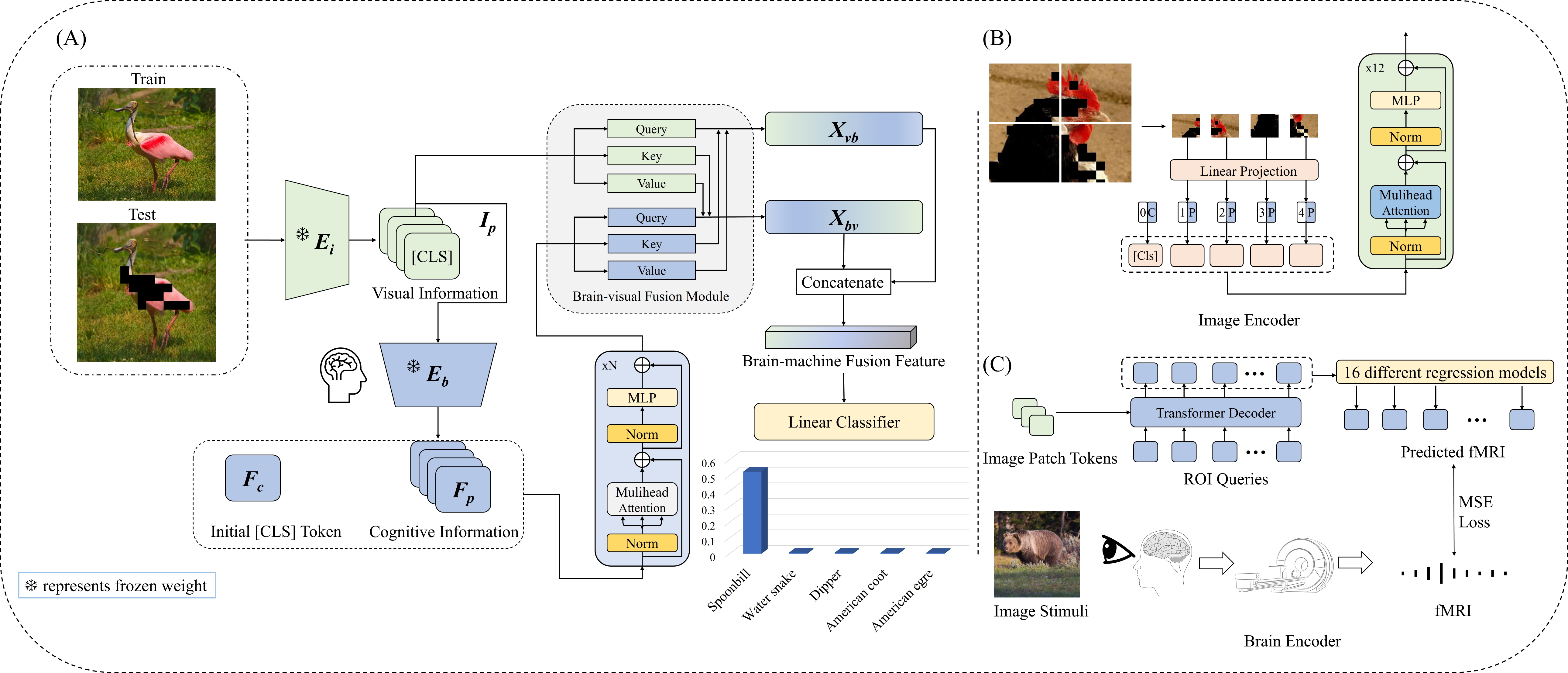}
	\caption{(A). Brain-machine Fusion model. Our learning process can be divided into four steps: image feature extraction, generation of brain fMRI via image features, fMRI feature extraction and multimodal modeling. The fused multimodal representation is used to predict the corresponding category via a linear classifier. (B). The frozen weighted DINOv2 model is applied as the image encoder, from which image \texttt{[CLS]} token and patched image tokens are extracted. (C). The training process and architecture of the brain encoder. The NSD dataset is used to train the model, the image patch tokens are extracted from the image encoder with the input of the NSD image stimuli. The brain encoder is optimized with mean square error (MSE) loss calculated between the predicted fMRI and the raw fMRI data.}
	\label{fig:new}
\end{figure*}

We propose a novel BMFL framework to combine cognitive information (i.e., human cerebrum prior knowledge) with visual information, improving the OOD generalization ability of traditional CV models. Fig. \ref{fig:new}(A) is the overview of the BMFL framework, which contains four fundamental modules: image encoder, brain encoder, brain transformer and brain-visual fusion module. In order to obtain better OOD generalization performance, we need to choose a reliable CV backbone first. We choose the self-supervised models as the image encoder as it can learn category-independent features with rich semantic information, which can extract robust image features from the input images. Regarding the brain encoder, it can predict the brain visual fMRI from the image features. After obtaining the fMRI data, two main issues need to be addressed: how to extract and utilize the prior cognitive knowledge from the fMRI. To handle these two issues, the brain transformer is proposed, and the cross-attention mechanism is used for integrating the visual features with the cognitive knowledge extracted from the brain transformer module.

\subsection{Image Encoder}
In the proposed BMFL framework, DINOv2 (ViT-b) \cite{dinov2} is used for the image encoder $\bm{E_i}$. Fig. \ref{fig:new}(B) shows the architecture of the image encoder. DINOv2 adopts self-distillation method to learn the semantic features from the input images. Specifically, researchers conduct a unsupervised pre-training method by minimizing the outputs distance between the large parameter teacher model and fewer parameter student model. The experimental results shows that using self-distillation method can obtain higher performance than training a ViT model from scratch. The sophisticated design and high-quality large-scale pre-training dataset fortify the robustness of the image encoder. Without extra fine-tuning, DINOv2 model can obtains the equivalent performance to the supervised models in the CV downstream tasks. The image encoding process is calculated as:
\begin{equation}
	\label{eq:img}
	[\bm{I_c}, \bm{I_p}] = \bm{E_i}(\bm{I})
\end{equation}
where $\bm{E_i}(\cdot)$ represents the image encoder, $\bm{I_c}$ and $\bm{I_p}$ are the image \verb+[CLS]+ token and patch token extracted from the image encoder respectively. $\bm{I_p}$ is then used for the input of the brain encoder $\bm {E_b}$ to get the corresponding brain visual fMRI data. 

\subsection{Brain Encoder}
The brain visual fMRI plays a crucial role in the proposed BMFL framework. Conventionally, the brain visual fMRI is acquired by presenting visual stimuli to the subjects and recording the blood-oxygenation-level-dependent (BOLD) responses. However, the fMRI data collection is expensive and time-consuming, so it's impossible for us to collect and process millions of brain visual fMRI. In order to handle this problem, we employ a pre-trained DETR \cite{detr} brain visual fMRI prediction model \cite{adeli2023predicting} as the brain encoder $\bm {E_b}$, which is pre-trained on the NSD Dataset \cite{nsd}, the best quality open-source brain dataset to date. The 7-Tesla scanner provides high-quality and low-noise brain visual fMRI data, which enhances the reliability of the brain encoder. 

In the pre-training process, Andeli \emph{et al.} utilize the DINOv2 as its image encoder. After the image encoding process, the patch image tokens $\bm{I_p}$ are served as the key and the value of the transformer decoder \cite{attention}. The input queries of the decoder correspond to different brain regions of interest (ROIs) from different hemispheres \cite{adeli2023predicting}. Fig. \ref{fig:new}(C) shows the brain encoder architecture and its pre-training process. The decoding output tokens represent eight stream level ROIs: early, mid-ventral, mid-lateral, mid-parietal, ventral, lateral, parietal and unknown (all the unsigned brain vertices) for each left and right hemisphere. The decoding output tokens are then aligned with the NSD fMRI data through a linear regression model, updating the parameter by minimizing the MSE loss.

Given the patch image tokens $\bm{I_p}$, the predicting process of fMRI follows:
\begin{equation}
	\label{eq:brainencoder}
	\bm{fMRI} = \bm{E_b}(\bm{I_p})
\end{equation}
where $\bm{E_b}(\cdot)$ represents the pre-trained brain encoder model.

To verify the reliability of the predicted fMRI, we visualize the correlation between the predicted fMRI and the ground truth fMRI by PyCortex \footnote{\url{https://github.com/gallantlab/pycortex}} tool. The evaluation metric $m$ can be calculated as:
\begin{equation}
	m = mean\{\frac{R_1^2}{NC_1},\frac{R_2^2}{NC_2},…,\frac{R_n^2}{NC_n}\}*100
\end{equation}
where n denotes the subject number, NC denotes the noise ceiling, R is the Pearson correlation coefficient (PCC) calculated follows:
\begin{equation}
	\label{eq:pcc}
	R =\frac{\sum\limits_{i=1}^n\left(\left(x_i-\bar{x}\right)\left(y_i-\bar{y}\right)\right)}{\sqrt{\sum\limits_{i=1}^n\left(x_i-\bar{x}\right)^2}\sqrt{\sum\limits_{i=1}^n\left(y_i-\bar{y}\right)^2}}
\end{equation}
where $x$ denotes the ground truth fMRI, $\bar{x}$ is the average value of $x$, $y$ denotes the predicted fMRI, $\bar{y}$ is the average value of $y$. The brain encoder reached a score of 63.5229 $m$ on the Algonauts2023 \cite{2023algonauts} released test datasets.

\begin{figure}[htbp!]
	\centering
	\includegraphics[width=\linewidth]{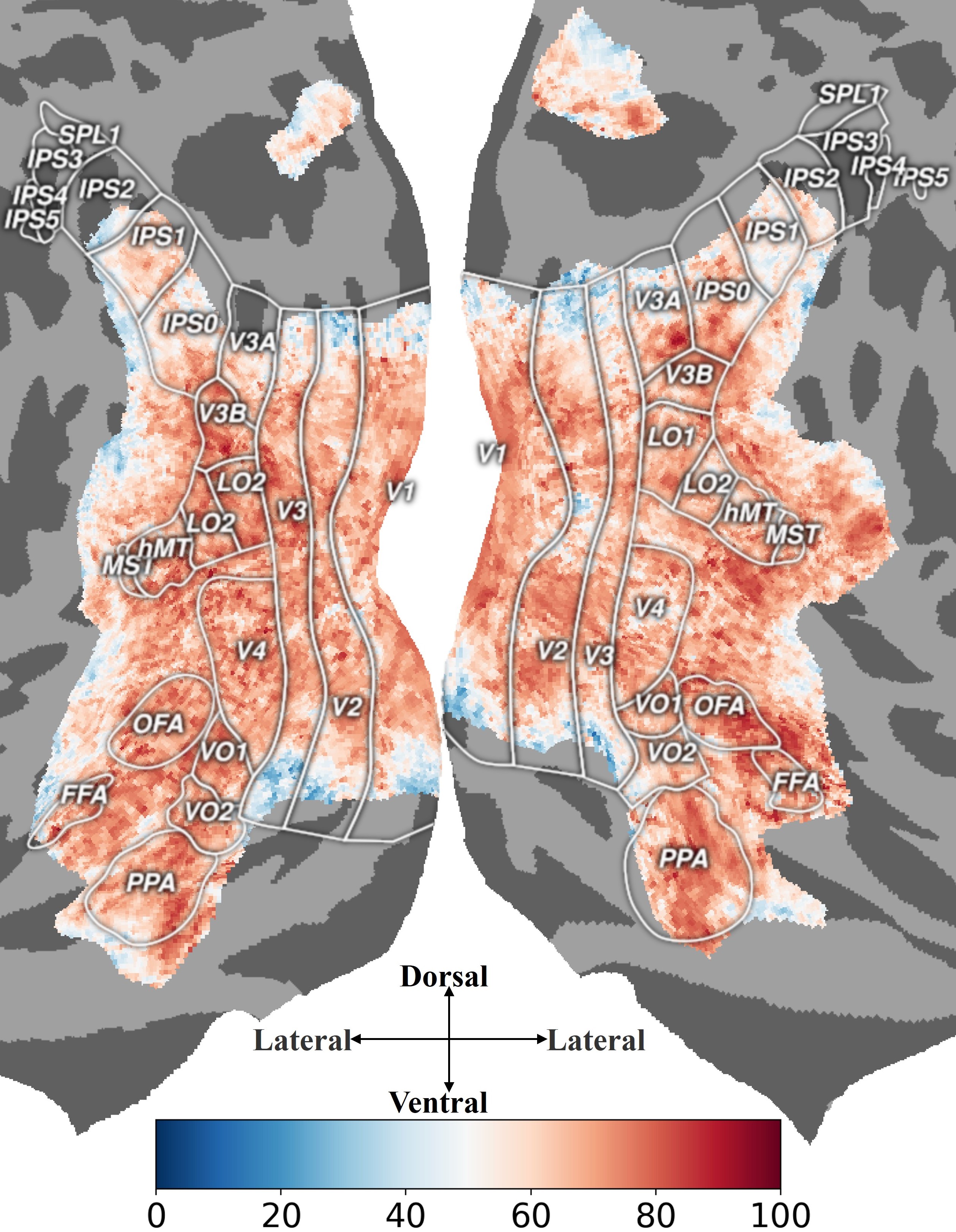}
	\caption{The reported result from \cite{adeli2023predicting} of the correlation between the predicted fMRI and the ground truth fMRI.}
	\label{fig:pycortex}
\end{figure}

\subsection{Brain Transformer}
After obtaining the fMRI, we should then consider how to extract knowledge from them. Previous works \cite{du2023decoding,DuTNNLS} usually use a simple linear model to extract knowledge from the fMRI. Although these methods perform well on visual neural decoding tasks, there are still several limitations. Firstly, the representation capability of the simple linear model is inferior than the transformer architecture. Besides, the main issue is the dimension discrepancy between fMRI features and transformer tokens. As the unified architecture can make it easier to build a multimodal learning model. We propose a brain transformer to learn the hierarchical knowledge from brain visual fMRI with low feature dimension that aligned to the transformer architecture. 

\begin{figure*}[htbp]
	\centering
	\includegraphics[width=\linewidth]{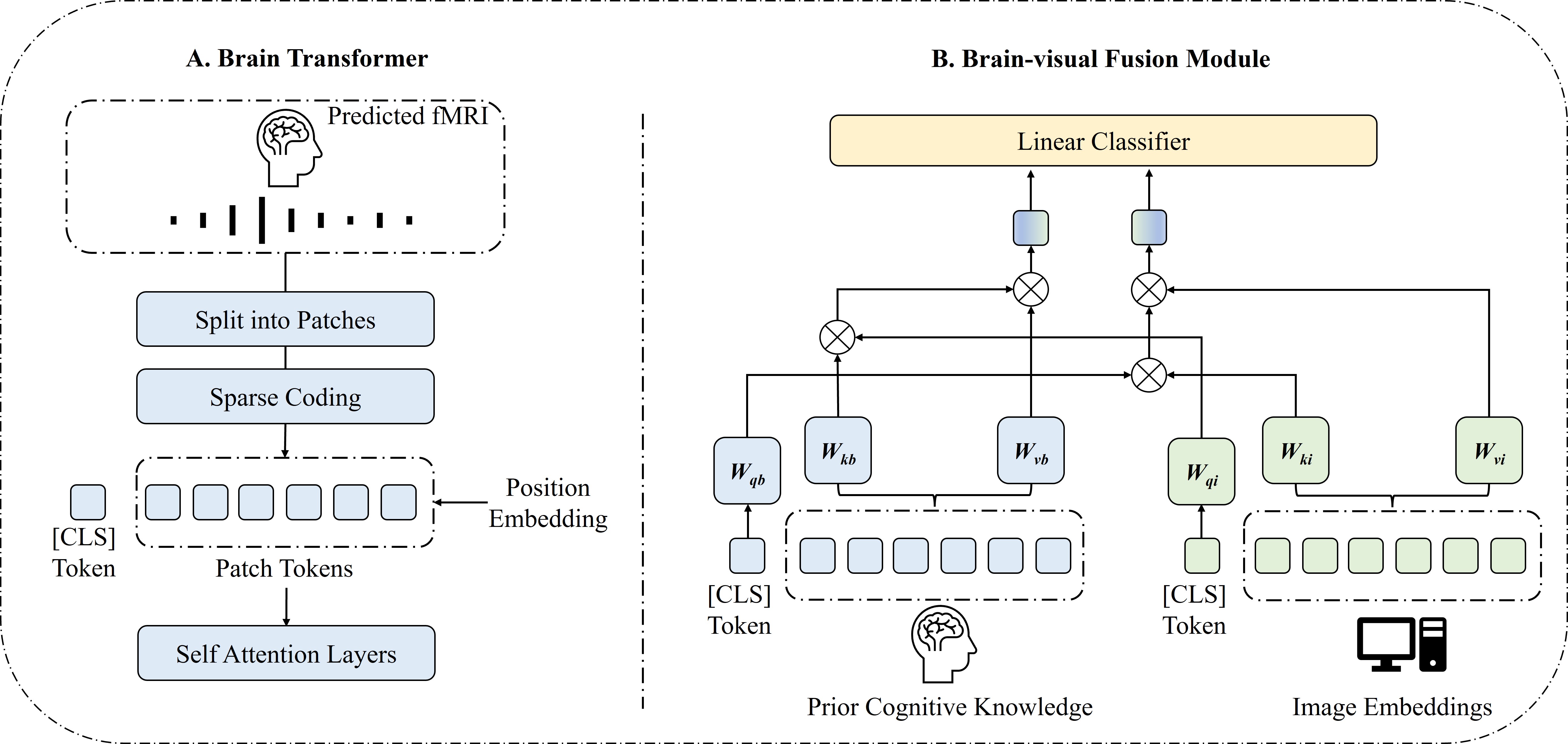}
	\caption{(A). The fMRI feature extractor.Inspired by ViT, we propose the brain transformer to extract knowledge from fMRI data. We slice the fMRI into several patches using convolutional operations, then add learnable position embeddings along with the \texttt{[CLS]} token. These fMRI patch tokens and \texttt{[CLS]} token are the input transformer encoder layer. (B). Brain-visual fusion module. The cross-attention mechanism has proven its strong capability in multimodal fusion in generative models. In the proposed BMFL framework, the \texttt{[CLS]} token is used as query of the cross-attention, while another modal's patch tokens are used as the key and value. The fused features are then concatenated with the global average pooling features of image patches to form the input for the linear classifier.}
	\label{fig:braintransformer}
\end{figure*}

In order to mitigate the adverse effects of similar response values between neighboring voxels to the model, we conduct a sparse code modeling method. In the proposed brain transformer, the predicted fMRI is divided into patches by 1D convolution operator via the sparse coding method \cite{mindvis}. Specifically, the kernel size and the stride of the 1D convolution operator are 192 which is equivalent to the embedded dimension of ViT tiny. The number of output channels is 768 that equivalent to the embedded dimension of ViT base. The 192-dimensional fMRI is projected into 768 dimensions by sparse coding, which reduces the influence of adjacent voxel responses on the model and maintains a considerable sequence length. Then, we add the position embedding and the \verb+[CLS]+ token respectively. The \verb+[CLS]+ token $\bm{F_c}$ will gradually learn the global features of the fMRI follows:

\begin{equation}
	\label{eq:braintrans}
	[\bm{F_c}, \bm{F_p}] = \bm{E_f}(\bm{fMRI}),
\end{equation}
where $\bm{E_f}(\cdot)$ is the proposed brain transformer. Consequently, we obtain prior cognitive knowledge encapsulated in low-dimensional representations derived from predicted brain fMRI. Fig. \ref{fig:braintransformer}(A) shows the proposed brain transformer architecture.

\subsection{Brain-visual Fusion Module}
Following the image encoding, fMRI predicting and the fMRI feature extraction stages, we propose a Brain-visual Fusion (BVF) module to learn the joint representation of the fMRI features and visual features $\bm{X_j}$ via the cross-attention mechanism \cite{crossvit}. In the proposed BVF module, the extracted fMRI's \verb+[CLS]+ token $\bm{F_c}$, and patch tokens $\bm{F_p}$, in conjunction with the visual features $\bm{I_c}$ and $\bm{I_p}$ are served as the key and value, or alternatively, the query of cross-attention. Eq. \ref{cross} describes the fusion process in the proposed BVF module. Through the BVF module, DNNs are capable of effectively fusing the prior cognitive knowledge from the human brain with visual features. Letting prior cognitive knowledge to mitigate the domain shift phenomenon of DNNs. Fig. \ref{fig:braintransformer}(B) shows the architecture of the BVF method.
\begin{equation}
	\label{cross}
	CrossAttention(Q,K,V) = softmax(\frac{QK^T}{\sqrt{d_k}})V
\end{equation}

More specifically,
\begin{equation}
	X_{vb} = CrossAttention(\bm{I_c}, \bm{F_p}, \bm{F_p})
\end{equation}		
\begin{equation}
	X_{bv} = CrossAttention(\bm{F_c}, \bm{I_p}, \bm{I_p})
\end{equation}
where $X_{vb}$ and $X_{bv}$ represents the image-query multimodal feature and the brain-query multimodal feature respectively. Moreover, we fuse these features by:
\begin{equation}
	\label{eq2}
	X_j = Concatenate(X_{vb}, X_{bv})
\end{equation}
where $X_j$ is the brain-machine joint modeling feature. 

\subsection{Pearson Correlation Coefficient Maximization}
In the multimodal learning research, reducing the distance between different modalities' representations is one of the key priorities. Some researchers utilize constructive loss \cite{clip,blip,blip2}, while others use distance metrics to measure and thus reduce the distance between different modalities' representations \cite{liu2023brain,du2023decoding}. In the proposed BMFL framework, we optimize the fusion process by maximizing the PCC $R$ value. Minimizing the discrepancy between two modalities enhances the fusion capability, and thus improving the OOD generalization ability.

The $R$ value, as shown in Eq. \ref{eq:pcc}, is one of the most important indicators of the degree of the linear correlation between two variables. Specifically, $R$=1 indicates the perfect positive linear correlation between the two variables, $R$=0 indicates the two variables are linearly independent, while $R$=-1 indicates the perfect negative linear correlation of the two variables. In Eq. \ref{eq:pcc}, $x$ denotes the fusion feature $x_{vb}$, $\bar{x}$ is the average value of $x$, $y$ denotes the another fusion feature $x_{bv}$, $\bar{y}$ is the average value of $y$.

\begin{algorithm}[htbp]
	\caption{BMFL framework}
	\textbf{Training}\\
	\label{alg:algorithm}
	\textbf{Input}: Image $I$
	\begin{algorithmic}[1] %[1] enables line numbers
		\FOR{epoch in training epochs}
		\STATE $\bm{I_c}, \bm{I_p} \leftarrow \bm{E_i}(I)$
		\STATE $fMRI \leftarrow \bm{E_b}(\bm{I_p})$
		\STATE $\bm{F_c}, \bm{F_p} \leftarrow \bm{E_f}(fMRI)$
		\STATE $\bm{X_{vb}} = CrossAttention(\bm{I_c}, \bm{F_p}, \bm{F_p})$
		\STATE $\bm{X_{bv}} = CrossAttention(\bm{F_c}, \bm{I_p}, \bm{I_p})$
		\STATE $\bm{X_j} = Concatenate(\bm{X_{vb}}, \bm{X_{bv}})$
		\STATE Categorical Probability $\leftarrow MLP(\bm{X_j})$
		\STATE Compute $\mathcal L_{cls}$ and $\mathcal L_{fusion}$ 
		\STATE backpropagation
		\ENDFOR
		\STATE \textbf{return} Categorical Probability
	\end{algorithmic}
	
	\textbf{Testing}\\
	\textbf{Input}: OOD Image $\hat{I}$
	\begin{algorithmic}[1] %[1] enables line numbers
		\STATE $\bm{\hat{I}_c}, \bm{\hat{I}_p} \leftarrow \bm{E_i}(\hat{I})$
		\STATE $fMRI_{OOD} \leftarrow \bm{E_b}(\bm{\hat{I}_p})$
		\STATE $\bm{\hat{F}_c}, \bm{\hat{F}_p} \leftarrow \bm{E_f}(fMRI_{OOD})$
		\STATE $\bm{X_{vb}} = CrossAttention(\bm{\hat{I}_c}, \bm{\hat{F}_p}, \bm{\hat{F}_p})$
		\STATE $\bm{\hat{X}_{bv}} = CrossAttention(\bm{\hat{F}_c}, \bm{\hat{I}_p}, \bm{\hat{I}_p})$
		\STATE $\bm{\hat{X}_j} = Concatenate(\bm{\hat{X}_{vb}}, \bm{\hat{X}_{bv}})$
		\STATE Categorical Probability $\leftarrow MLP(\bm{\hat{X}_j})$
		\STATE \textbf{return} Categorical Probability
	\end{algorithmic}
	
	\textbf{Output}: Categorical Probability
\end{algorithm}

\subsection{Overall loss function and training process}
In the proposed BMFL framework, $\mathcal L_{BMFL}$ \ref{eq:loss_task} is used to regulate the model. 
\begin{equation}
	\centering
	\label{eq:loss_task}
	\mathcal L_{BMFL} = \mathcal{L}_{cls} + \alpha \mathcal{L}_{fuion}
\end{equation}
where $\mathcal{L}_{cls}$ \ref{eq:CE} is a standard cross-entropy loss, while $\mathcal{L}_{fuion}$ is a non-iterative PCC calculation method following Eq. \ref{eq:fusion}. The hyper-parameter $\alpha$ is the regularization weight that dictates the degree of contribution of the fusion regularization $\mathcal{L}_{fuion}$ to $\mathcal L_{BMFL}$.
\begin{equation}
	\centering
	\label{eq:CE}
	\mathcal{L}_{cls}(y,\bar{y}) = -\sum_{i=1}^{n}y(i)\log \bar{y}(i)
\end{equation}
\begin{equation}
	\centering
	\label{eq:fusion}
	\mathcal{L}_{fusion}(x,y) = \frac{n\sum\limits_{i=1}^n x_i y_i - (\sum\limits_{i=1}^n x_i)(\sum\limits_{i=1}^n y_i)}{\sqrt{n\sum\limits_{i=1}^n x_i^2 - (\sum\limits_{i=1}^n x_i)^2n\sum\limits_{i=1}^n y_i^2 - (\sum\limits_{i=1}^n y_i)^2}} 
\end{equation}
where $y$ is the true probability distribution, $\bar{y}$ is the predicted probability distribution. We optimize the proposed model by minimizing $\mathcal{L}_{BMFL}$, while the component $\mathcal{L}_{fusion}$ must be maximized to yield effective results. Thus the regularization weight $\alpha \in (-\infty,0]$, where $\alpha=0$ means the $\mathcal L_{BMFL}$ is $\mathcal{L}_{cls}$.

Algorithm \ref{alg:algorithm} shows the PyTorch pseudocode for the implementation of the proposed BMFL framework.

\section{Experiments}
\subsection{Datasets}
\begin{figure*}[htbp!]
	\centering
	\includegraphics[width=\linewidth]{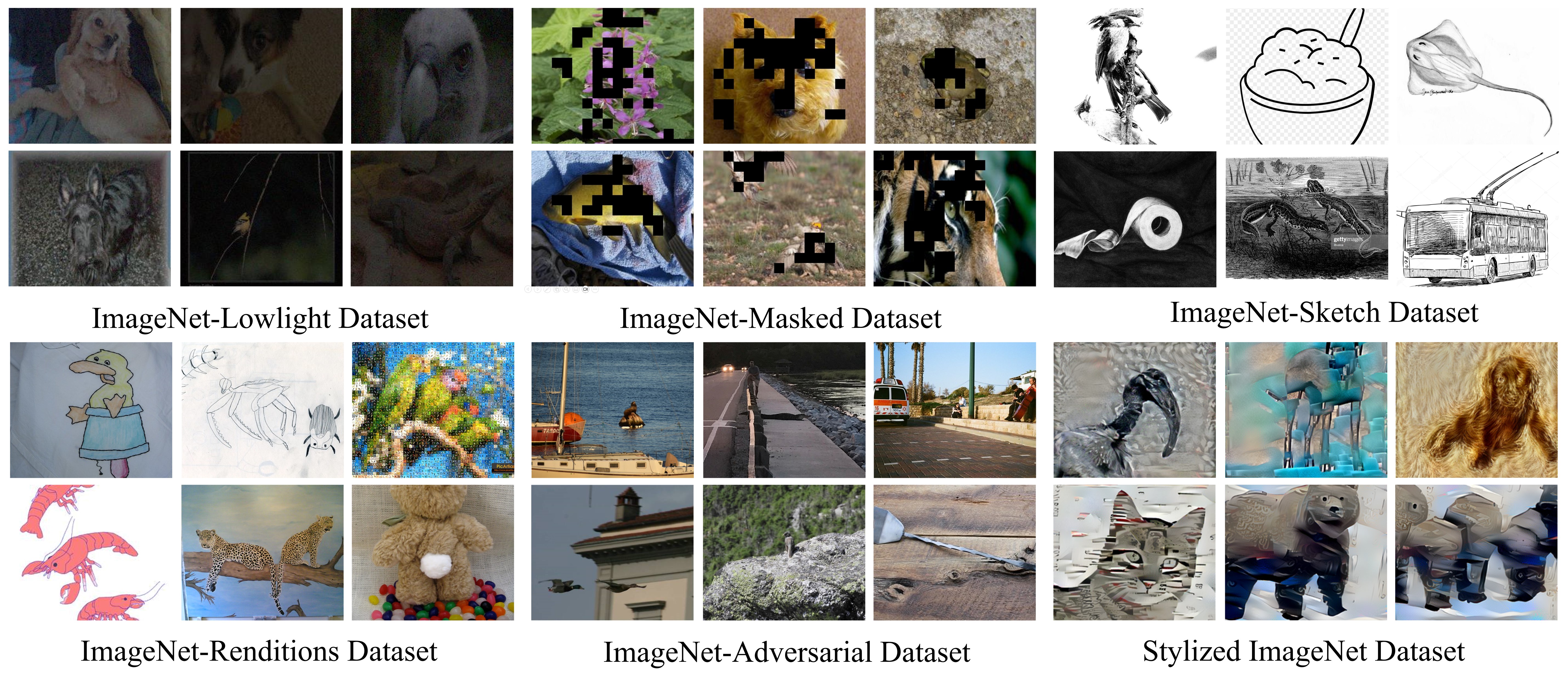}
	\caption{Two self-created and four open-source OOD datasets based on the ImageNet-1K dataset.}
	\label{fig:dataset}
\end{figure*}
\textbf{Training dataset}. The ImageNet-1k dataset is one of the most famous image classification dataset with 1000 different categories. In the following experiments, we adopt the ImageNet-1k training dataset during the training process.

\textbf{Testing datasets.} There are a total of 7 datasets that are used to test the performance of our BMFL framework. Within these 7 datasets, the ImageNet-1k validation dataset is used to test the ID classification performance and the remaining 6 datasets are used for the OOD generalization testing. To make the BMFL framework more theoretical and applied, we use 4 mainstream open-source OOD datasets and create 2 OOD datasets to supply the real-world extreme OOD scenarios. Furthermore, the two self-created datasets and the Stylized ImageNet dataset are created based on the images of the ImageNet validation dataset, while the images of the other 3 open-source datasets are from different sources within the 1000 ImageNet labels.

{\bf{Stylized ImageNet dataset}} \cite{style} is a cross-domain "16-class-ImageNet" dataset, which contains 1280 images with 80 samples in each category. Researchers create the dataset by randomly selecting the ImageNet images first. The raw Stylized ImageNet dataset is not provided, but the researchers open-source the code for generating the images and we create the Stylized ImageNet dataset following \footnote{\url{https://github.com/rgeirhos/Stylized-ImageNet}}.

{\bf{ImageNet-Renditions}} (ImageNet-R) dataset \cite{hendrycks2021many} contains 30k image renditions with 200 ImageNet classes. Researchers first collect the images from Flickr website, then the collected images will be automatically filtered by the Amazon MTurk tools. In the end, the remaining images will be selected manually to ensure the quality of the dataset.

{\bf{ImageNet-Sketch}} (ImageNet-S) dataset \cite{wang2019learning} contains 50k sketch-like greyscale images with 1000 ImageNet classes (50 images per class). Researchers use Google Images website to collect these sketch-like greyscale images. Specially, for those classes that contain less than 50 images, researchers conduct a augmentation operation until the number of images reached to 50.

{\bf{ImageNet-Adversarial}} (ImageNet-A) dataset \cite{hendrycks2021natural} contains 7500 images with 200 ImageNet classes. The images are collected from 3 different sources: iNaturalist, Flickr and DuckDuckGo. After the adversarial filtering process, researchers chose the remaining images with low-confidence as the elements of the ImageNet-A dataset.

{\bf{Masked dataset}} is an ImageNet-based dataset. We create it in order to introduce one of the most common OOD samples in life (i.e., extreme scenarios) into the OOD generalization task. The self-supervised DINO model \cite{dino} is used for generating masked images through subject occlusion or using random occlusion to partially mask the foreground of the image.

{\bf{Low light dataset}} is another ImageNet-based dataset. Our motivation for producing this dataset is obvious: low light scenario is the most common OOD sample in the real-world applications,  as it appears everyday. The low light dataset was created through the steps of unprocessing procedure, low light corruption, and image signal processing following \cite{lowlight}.

\begin{table}[htbp]
	\centering
	\caption{Specific information about both in-domain and OOD datasets used for the experiments.}
	\resizebox{\linewidth}{!}{	
		\begin{tabular}{cccc}
			\hline
			Dataset & Amount & Categories & Distribution \\
			\hline
			\multicolumn{4}{c}{Self-created datasets}\\
			\hline
			Low light & 50000 & 1000 & OOD\\
			Masked & 50000 & 1000 & OOD\\
			\hline
			\multicolumn{4}{c}{Open-source datasets}\\
			\hline
			ImageNet-1k-val & 50000 & 1000 & $i.i.d$ with the training set\\
			ImageNet-Sketch & 50000 & 1000 &  OOD\\
			ImageNet-R & 30000 & 200 & OOD\\
			ImageNet-A & 7500 & 200 &  OOD\\
			Stylized ImageNet & 1280 & 16 & OOD\\
			\hline
			\end{tabular}
	}
	\label{tab:dataset}
\end{table}

\subsection{Implementation Detail}
In the baseline experiment, we use Timm \footnote{\url{https://github.com/huggingface/pytorch-image-models}} library to conduct the pre-training baseline models and compete the image classification performance on both ID and OOD datasets. 

During the training process, we use 512 brain-image data to optimize our BVL module in one iteration. The learning rate peaks at 5e-5 after 1.5 epochs with the linear warm-up and then decreases with a cosine learning rate schedule for the remaining epochs. We adopt -0.4 as the value of the regularization weight $\alpha$. Moreover, the AdamW optimizer \cite{shazeer2018adafactor} is employed to update the model parameters during the backpropagation process, with the hyper-parameters $\beta1$ = 0.9 and $\beta2$ = 0.999, and $\lambda$ = 0.02.

%\begin{table}[h]
%	\centering
%	\begin{tabular}{cc}
	%		\hline
	%		Hyper Parameters & Value \\
	%		\hline
	%		learning rate & 5e-5 \\
	%		epochs & 15 \\
	%		batch size & 512 \\
	%		weight decay & 0.001 \\
	%		\hline
	%	\end{tabular}
%	\caption{The specific hyper-parameter in our experiment.}
%	\label{tab::implemention detail}
%\end{table}

\subsection{Results}

\begin{table*}[htbp!]
	\centering
	\caption{OOD generalization test for different models. * denotes the brain-inspired model. The results highlighted in bold are the best.}
	\resizebox{\linewidth}{!}{
		\begin{tabular}{cccccccc}
		\hline
		Method & Val & Low light & Masked &  Stylized ImageNet & ImageNet-A &ImageNet-R & ImageNet-Sketch \\
		%		\cline{2-14}
		%		\multicolumn{1}{c}{Method} & top-1 & top-5 & top-1 & top-5 & top-1 & top-5 & top-1 & top-5 & top-1 & top-5 & top-1 & top-5 & top-1 & top-5 \\
		\hline
		\multicolumn{8}{c}{\bf{CNN Architecture}}\\
		\hline
		VGG16\cite{vgg} & 70.01\% & 35.41\% & 14.84\% & 11.56\% & 3.61\% & 27.01\% & 16.94\%\\
		CORnet-S*\cite{KubiliusSchrimpf2019CORnet} & 73.02\% & 39.98\% & 17.70\% & 18.83\% & 0.84\% & 34.00\% & 21.56\%\\
		Inception-v4\cite{inception}  & 79.41\%  & 58.24\% & 27.41\% & 20.94\% & 13.25\% & 41.65\% & 32.85\%\\
		EfficientNet-B0\cite{efficientnet} & 76.20\% & 52.51\% & 28.27\% & 21.02\% & 7.24\% & 35.68\% & 24.99\%\\
		ResNet-50\cite{ResNet} & 79.38\% & 60.01\% & 56.51\% & 17.34\% & 9.81\% & 39.59\% & 29.75\% \\
		\hline
		\multicolumn{8}{c}{\bf{ViT Architecture}}\\
		\hline
		ViT-b\cite{vit} & 84.71\% & 75.05\% & 68.72\% & 37.03\% & 41.59\% & 58.77\% & 45.10\% \\
		DeiT III (ViT-b, ImageNet-1k)\cite{deit3} & 81.40\% & 73.22\% & 65.63\% & 35.55\% & 31.97\% & 53.84\% & 41.33\% \\
		DeiT III (ViT-b, ImageNet-21k)\cite{deit3} & 84.79\% & 73.62\% & 66.24\% & 32.42\% & 49.63\% & 60.06\% & 46.82\% \\
		DINOv2 (ViT-b, backbone)\cite{dinov2} & 84.50\% & 74.95\%  & 71.60\% & 36.56\% & 53.65\% & 67.31\% & 53.66\% \\
		\hline
		{\bf{Ours}}& {\bf{85.72\%}} & {\bf{76.56\%}} &  {\bf{72.70\%}} & {\bf{37.34\%}} & {\bf{57.04\%}}& {\bf{68.79\%}}& {\bf{55.40\%}}\\
		%		\textbf{Ours-l2(wd),$\alpha$=0.5}&85.76\% &97.67\%&76.17\%&93.18\%&71.79\%&90.33\%&37.19\%&62.97\%\\
		%		\textbf{Ours-l2,$\alpha$=0.5}&85.76\% &97.68\%&76.08\%&93.14\%&71.92\%&90.31\%&36.95\%&63.67\%\\
		%		\textbf{Ours-l2,$\alpha$=1}&85.70\% &97.71\%&76.25\%&93.33\%&72.05\%&90.35\%&37.03\%&64.22\%\\
		\hline

		\end{tabular}
	}
	\label{tab:exp}
\end{table*}

We compare the OOD generalization ability of different model architectures, including CNNs such as VGG16 \cite{vgg}, Inception-v4 \cite{inception}, ResNet-50 \cite{ResNet} and EfficientNet-B0 \cite{efficientnet}. ViT-based models like DeiT III \cite{deit3}, the image encoder backbone DINOv2 \cite{dinov2} and the brain-inspired model CORnet-S \cite{KubiliusSchrimpf2019CORnet}, our BMFL framework demonstrates superior OOD generalization performance among these baseline models. As illustrates in Table \ref{tab:exp}, ViT architecture exhibits stronger OOD generalization capability than CNN models. This experimental phenomenon can be attributed to two main factors. The first one is the designing of the transformer architecture (i.e., ViT-based models) can better focus on global features through the self-attention mechanism, whereas CNN models predominately rely on the local features extracted by convolutional layers. Secondly, the scale of the pre-training dataset plays a crucial role in the OOD generalization capabilities. In general, ViT-based models that have large parameters can fit large-scale datasets easily, while CNN models are struggle with it. Obviously, high-quality large-scale datasets can provide more information with few noise that can help DNNs to focus on those meaningful information, thus improving the knowledge transfer ability and the OOD generalization capability. In our experiment, we compare the ID and OOD generalization performance of the same model with pre-training datasets. We utilize the DeiT III (ViT-b) model as a representative example. Compared to the ImageNet-1k pre-training dataset, the model pre-trained on the large-scale ImageNet-21k dataset and fine-tuned on the ImageNet-1k datasets exhibits superior OOD generalization capability in most OOD datasets, with the exception of the Stylized ImageNet dataset. The image encoder backbone DINOv2 is another convincing evidence, the large-scale high-quality pre-training dataset LVD-142M is another key factor in achieving such results besides the self-supervised design. Furthermore, the brain-inspired CORnet-S model achieves and even surpasses some DNNs with its shallow model layers. This experimental result shows that conducting a brain-inspired model is a effective way to obtain high OOD generalization capability with few parameters.

Different learning methods can also influence the OOD generalization performance. Compared to the label-driven supervised methods, self-supervised models can learn more semantic information to mitigate the domain shift phenomenon.  As a common representation learning method, the dominant approach in OOD generalization task, conducting a self-supervised model is another way to obtain the high OOD generalization performance. Additionally, we observed that if the domain distance is too large, the large-scale pre-training dataset brings a negative impact to the label-driven supervised models.

The proposed BMFL framework obtains the highest OOD generalization performance, the key of its success can be summarized into two reasons: 1. The effect of the prior cognitive knowledge. 2. The advantage of the cross-attention mechanism and the PCC maximization regulation method. We conduct the following ablation studies to explore how the aforementioned factors influence the OOD generalization capability.

\subsection{Ablation Study}
We conduct the first ablation study to investigate how different fusion methods affect the OOD generalization ability, the results are shown in Table \ref{tab:ablation}. We eliminate the cross-attention module and the l2 loss $L_{fusion}$, using a simple concatenation fusion method to combine the image \texttt{[CLS]} token $\bm{I_c}$ and the fMRI \texttt{[CLS]} token $\bm{F_c}$ directly as the classification features. Our completed BMFL framework demonstrates the best OOD generalization performance on the Low light and Masked OOD datasets. Comparing with the visual backbone DINOv2, introducing fMRI directly fails to bring the improvement and even reduces the OOD generalization capability. Comparing with the concatenation fusion method, the cross-attention mechanism and PCC maximization method demonstrate the advantages on the brain-machine fusion task respectively, successfully improving the OOD generalization capability of the CV backbone. In particular, it is important to note that the Stylized ImageNet dataset is the most distinctive one within the three datasets, the more complex the multimodal fusion method is, the worse OOD generalization ability the model has. In conjunction with the aforementioned brain encoder method, the fMRI is predicted with the input features of the visual backbone DINOv2. Due to the low classification accuracy of the Stylized ImageNet dataset, the quality of the corresponding predicted fMRI are thereby influenced. Considering the high noise contained in the predicted fMRI, the lightweight fusion method will not damage the image features as well as the fMRI features during the fusion process and obtains a higher classification performance. 

\begin{table}[htbp!]
	\centering
	\caption{The classification accuracy between different fusion methods (concatenation or cross-attention) and learning methods (visual only or BMFL) on the in-domain and OOD datasets. Bold numbers indicate the best performance in each dataset.}
	\resizebox{\linewidth}{!}{
		\begin{tabular}{ccccc}
			\hline
			& & \multicolumn{3}{c}{top-1 accuracy} \\
			model & Variants & Low light & Masked & Stylized ImageNet\\
			\hline
			BMFL & $w/o$ fMRI & 74.95\% & 71.60\% & 36.56\% \\
			BMFL & $w/o$ cross-attention & 74.48\% & 70.06\% & {\bf{38.44\%}} \\
			BMFL & $w/o$ $\mathcal{L}_{fusion}$ & 75.99\% & 71.89\% & 37.58\% \\
			\hline
			{\bf{BMFL (Ours)}}& Completed & {\bf{76.56\%}} & {\bf{72.70\%}} &  37.34\% \\
			\hline
		\end{tabular}
	}
	\label{tab:ablation}
\end{table}

%\begin{figure*}[htbp!]
%	\centering
%	\includegraphics[width=\linewidth]{confusion matrix}
%	\caption{Confusion matrices for the OOD generalization ability of three different methods on the stylized dataset.}
%	\label{fig:cm}
%\end{figure*}

In the second ablation experiment, we examine how different brain ROIs affect the OOD generalization performance. According to the previous works, the brain visual cortex can be simply divided into two portions, the low-level visual cortex (LVC) and the high-level visual cortex (HVC). In our experiment, we follow the division of \cite{2023algonauts}, where LVC contains four ROIs: V1, V2, V3 and hV4. As for HVC, the ventral, lateral and parietal cortex can be categorized into the HVC area. In general, the LVC in the occipital region contains contour and color information, while the HVC contains more semantic information \cite{neuron}.

\begin{table}[h]
	\centering
	\caption{Comparison of OOD generalization ability with LVC, HVC and all brain visual cortex. Bold numbers indicate the best performance in each dataset.}
	\resizebox{\linewidth}{!}
	{\begin{tabular}{ccccc}
			\hline
			& & \multicolumn{3}{c}{top-1 accuracy} \\
			model & variants & Low light & Masked & Stylized ImageNet \\
			\hline
			BMFL & $w/o$ fMRI & 74.95\% & 71.60\% & 36.56\% \\
			BMFL & LVC & 76.42\% & 72.25\% & 37.11\% \\
			%		BMF & HVC & 76.04\% & 70.07\% & 38.67\% \\
			%		\hline
			BMFL & LVC+HVC & {\bf{76.56\%}} & {\bf{72.70\%}} &  {\bf{37.34}}\% \\
			\hline
		\end{tabular}
	}
	\label{tab:ROI}
\end{table}
As shown in Table \ref{tab:ROI}, even if we only use the LVC region, we can still effectively improve the OOD generalization ability comparing with the DINOv2 backbone. While using both LVC and HVC can balance the semantic and detailed information, thereby obtaining a greater performance. The experimental results can be corroborated with the two-streams hypothesis \cite{hickok2007cortical} of visual processing mechanism. Using both LVC and HVC can provide auxiliary information from both 'where' and 'what'. The results show that introducing the detail and semantic features can better guiding the perception process of CV model.

\section{Conclusion}
In this paper, we present a novel brain-machine fusion learning framework as well as two self-created OOD classification datasets based on ImageNet-1k validation dataset. Through the above experiments, we have confirmed that introducing prior cognitive  knowledge from human brain into DNNs can effectively mitigate the domain shift phenomenon. Letting prior cognitive knowledge guide the perception process of CV model can improve the OOD generalization ability. Through the ablation study, we compared how different fusion methods and different brain regions affect the OOD generalization performance of the proposed BMFL framework. The experimental results are consistent with the basic assumptions of both CV as well as cognitive neuroscience. 

In our future work, we are willing to develop a robust visual neural encoding model with the lightweight architecture, reducing the computational requirements while improving the quality of predicted fMRI.

\section*{Acknowledgements}
This work was supported in part by the National Key R\&D Program of China 2022ZD0116500; in part by the National Natural Science Foundation of China under Grant 62206284; in part by Beijing Natural Science Foundation under Grant L243016; and in part by Faculty of Education, Capital Normal University-Huawei Group Digital Intelligent Teacher Education Special.

\nocite{langley00}

\bibliography{shuangchenzhao}
\bibliographystyle{icml2024}

%%%%%%%%%%%%%%%%%%%%%%%%%%%%%%%%%%%%%%%%%%%%%%%%%%%%%%%%%%%%%%%%%%%%%%%%%%%%%%%
%%%%%%%%%%%%%%%%%%%%%%%%%%%%%%%%%%%%%%%%%%%%%%%%%%%%%%%%%%%%%%%%%%%%%%%%%%%%%%%
% APPENDIX
%%%%%%%%%%%%%%%%%%%%%%%%%%%%%%%%%%%%%%%%%%%%%%%%%%%%%%%%%%%%%%%%%%%%%%%%%%%%%%%
%%%%%%%%%%%%%%%%%%%%%%%%%%%%%%%%%%%%%%%%%%%%%%%%%%%%%%%%%%%%%%%%%%%%%%%%%%%%%%%
%\newpage
%\appendix
%\onecolumn
%\section{You \emph{can} have an appendix here.}
%
%You can have as much text here as you want. The main body must be at most $8$ pages long.
%For the final version, one more page can be added.
%If you want, you can use an appendix like this one.  
%
%The $\mathtt{\backslash onecolumn}$ command above can be kept in place if you prefer a one-column appendix, or can be removed if you prefer a two-column appendix.  Apart from this possible change, the style (font size, spacing, margins, page numbering, etc.) should be kept the same as the main body.
%%%%%%%%%%%%%%%%%%%%%%%%%%%%%%%%%%%%%%%%%%%%%%%%%%%%%%%%%%%%%%%%%%%%%%%%%%%%%%%%
%%%%%%%%%%%%%%%%%%%%%%%%%%%%%%%%%%%%%%%%%%%%%%%%%%%%%%%%%%%%%%%%%%%%%%%%%%%%%%%%

\end{document}